\documentclass[letterpaper, 10 pt, conference]{ieeeconf}  % Comment this line out if you need a4paper

\IEEEoverridecommandlockouts                              % This command is only needed if 
\usepackage[table, dvipsnames]{xcolor}
\usepackage{amsmath} % DO NOT REMOVE THIS PACKAGE ! VERY  IMPORTANT
\usepackage{graphicx}
\usepackage{adjustbox}
\usepackage{multicol}
\usepackage{tabularx}
\usepackage{caption}
\usepackage{amssymb}
\usepackage{cellspace}
\usepackage{hyperref}
\usepackage{cite}
\usepackage{multirow}
\usepackage{float}

\def\BibTeX{{\rm B\kern-.05em{\sc i\kern-.025em b}\kern-.08em
    T\kern-.1667em\lower.7ex\hbox{E}\kern-.125emX}}

\usepackage[ruled,vlined]{algorithm2e}

\newcommand{\acro}{DA-VIL\xspace}

\title{\LARGE \bf
\acro: Adaptive Dual-Arm Manipulation with Reinforcement Learning and Variable Impedance Control
}

\author{Md Faizal Karim$^{1*}$, Shreya Bollimuntha$^{1*}$, Mohammed Saad Hashmi$^1$, Autrio Das$^1$, Gaurav Singh$^1$, \\ Srinath Sridhar$^3$, Arun Kumar Singh$^2$, Nagamanikandan Govindan$^1$, K Madhava Krishna$^1$
\thanks{*Equal Contribution}% <-this % stops a space
\thanks{$^{1}$Robotics Research Center, IIIT Hyderabad}
\thanks{$^{2}$University of Tartu, Estonia}
\thanks{$^{3}$Brown University}%
}

\begin{document}

\maketitle
\thispagestyle{empty} 
\pagestyle{empty}

%%%%%%%%%%%%%%%%%%%%%%%%%%%%%%%%%%%%%%%%%%%%%%%%%%%%%%%%%%%%%%%%%%%%%%%%%%%%%%%%
\begin{abstract}

Dual-arm manipulation is an area of growing interest in the robotics community. Enabling robots to perform tasks that require the coordinated use of two arms, is essential for complex manipulation tasks such as handling large objects, assembling components, and performing human-like interactions. However, achieving effective dual-arm manipulation is challenging due to the need for precise coordination, dynamic adaptability, and the ability to manage interaction forces between the arms and the objects being manipulated. We propose a novel pipeline that combines the advantages of policy learning based on environment feedback and gradient-based optimization to learn controller gains required for the control outputs. This allows the robotic system to dynamically modulate its impedance in response to task demands, ensuring stability and dexterity in dual-arm operations. We evaluate our pipeline on a trajectory-tracking task involving a variety of large, complex objects with different masses and geometries. The performance is then compared to three other established methods for controlling dual-arm robots, demonstrating superior results. Project page: 
\url{https://dualarmvil.github.io/Dual-Arm-VIL/}

% Keywords - Dual-arm manipulation, variable impedance control as optimization, reinforcement learning.

% We propose \acro, a novel framework integrates Reinforcement Learning with impedance control formulated as an optimization problem.
% \acro sets a new benchmark for dual-arm manipulation offering an adaptive solution for dual-arm tasks.
% Through this method, we aim to enhance the robot's ability to handle a wide range of complex tasks in diverse environments.

% \Madhav{I think that the abstract is not delivering the knockout punch. I think we need to say we propose a novel pipeline that combines advantages of policy learning based on environment feedback and gradient based optimization to learn controller gains as well as the control outputs. We show the efficacy of this framework through its superior performance vis a vis purely policy gradient or optimization based approaches for Dual Arm manipulation. To the best of our knowledge this is the first such work for Dual Arm manipulation showcasing precise tracking outputs for a variety of objects wit varying masses. }
\end{abstract}

% \GN{I think we should add what Prof Madhav said: we evaluate this pipeline on a trajectory-tracking task involving a variety of large, complex objects with different masses and geometries. The performance is then compared to three other established methods for controlling dual-arm robots, demonstrating superior results.}(done)

%%%%%%%%%%%%%%%%%%%%%%%%%%%%%%%%%%%%%%%%%%%%%%%%%%%%%%%%%%%%%%%%%%%%%%%%%%%%%%%%
\section{INTRODUCTION}

Robust grasping and dexterous manipulation are two of the most essential capabilities that a dual-arm robot is expected to possess. Dual-arm manipulation has applications ranging from industrial automation and service robotics to assistive technologies \cite{stabilize, trends_robot_manipulation, two_arms_better, cooperative_dualarm}. Unlike single-arm manipulation, dual-arm tasks involve complex interactions, such as grasping objects with both hands, coordinating motions to assemble parts, and handling objects requiring delicate and precise control. 
% Achieving precise coordination between two arms while managing dynamic forces and interactions during task execution is the primary challenge. The robot must not only control the position and movement of each arm but also regulate the forces exerted on the object and the interactive forces between the arms. 
% \GN{Suggestion: The terms position and movement - are you refering to grasped object's and arm's motion. If we want to squeeze introduction section, perhaps the last two sentences of 1st para, 2nd para, and some sentences of 3rd para could be combined as they highlight similar challenges}
% \setlength{\belowcaptionskip}{-10pt}
However, when two robotic arms simultaneously grasp a single object, a variety of constraints arise \cite{dualarm_survey} due to the close coupling between the arms and the object: \textbf{First}, the rigid object dynamically couples the two arms, meaning any force applied by one arm directly affects the object and, in turn, the other arm. \textbf{Second}, each arm's movement is limited by the other, reducing the overall workspace available for the task. 
\setlength{\belowcaptionskip}{-6pt}
\begin{figure}[h]
    \centering
    \captionsetup{font=footnotesize}
    \includegraphics[width=\linewidth]{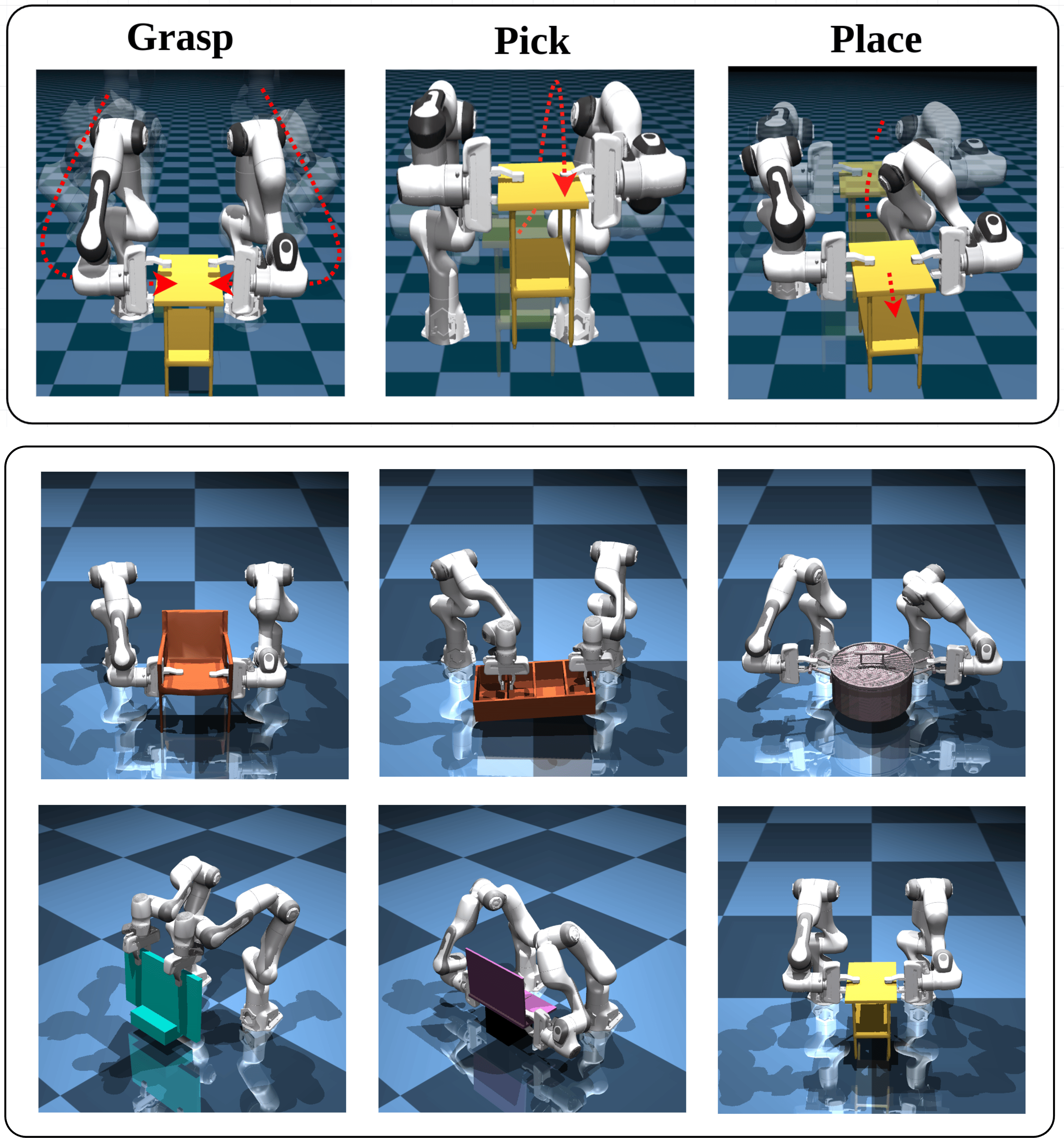}
    \caption{\textbf{Adaptive Dual-Arm Manipulation}. We present a novel framework that integrates Reinforcement Learning with an optimization-based Variable Impedance Control for efficient and adaptive dual-arm manipulation. Our approach handles a diverse set of objects varying in shape and mass. We define a pick-and-place task into three stages: \textbf{Grasp}, where the arm approaches the object and grasps it based on the input grasp poses, \textbf{Pick}, where the arms pick the object to an intermediate waypoint and \textbf{Place}, where the arms position the object at its goal pose.}
    \label{fig:enter-label}
\end{figure}
The entire trajectory must be within this shared workspace, which presents significant challenges for coordinated motion. \textbf{Third}, changes in object parameters—such as mass, geometry, etc. can generate destabilizing interaction forces that affect control performance. Designing controllers that account for these factors is essential to achieving coordinated and stable dual-arm manipulation.
This complexity highlights the need for compliant control strategies to manage the dynamic coupling between the arms, such as Impedance Control or Hybrid Torque control \cite{cartesian_impedance, dualarm_slabstone}. 
% In the absence of compliant control, rigid coupling can lead to significant stresses on the grasped object due to large forces and torques. 
% Furthermore, each arm is constrained by the other’s movement, and dynamic forces may destabilize the system. 
% Unlike traditional high-gain error feedback controllers, which dominate classical robotics 
% \Madhav{may want to cite here and at places similar where we make these broad statements} \cite{2}, impedance control offers a more feasible solution by dynamically adjusting stiffness and damping to accommodate uncertainties in position and large interaction forces during manipulation tasks. 
This adaptability allows the dual-arm system to respond effectively to dynamic forces, reducing tracking errors and enhancing performance. 
However, traditional control approaches like impedance control~\cite{impedance_control_original} require careful parameter tuning \cite{abu_frontier_vic, vices, self_tuning_ic}. 
Factors such as object geometry, mass, robot configuration, and task-specific trajectories significantly affect the choice of control parameters. Since these parameters are task-dependent and vary in real-time, a control approach with fixed parameters is often insufficient \cite{dualarm_survey}.

%\GN{the term static control may not be the appropriate- instead can we say a controller with fixed parameters? or controller parameters should adapt to changing conditions} Establishing a dynamic relationship between control variables, such as the object's pose and the wrenches applied by the arms, is critical to enhancing versatility and stability. A more adaptable control strategy—such as Variable Impedance Control (VIC) \Madhav{cite} with dynamically adjustable parameters—becomes essential for achieving both stability and dexterity in bimanual operations.
% However, traditional VIC methods require careful tuning and do not adapt well to changing environments or object parameters, limiting their effectiveness. Pure reinforcement learning (RL) approaches, while theoretically capable of learning such adaptations, face significant limitations. RL systems are often sample-inefficient, costly to train, and struggle with high-dimensional action spaces, making them less desirable for real-world bimanual manipulation tasks \cite{bidex, bimanual_sim2real}. 

To overcome these limitations, we propose an integrated approach that combines \textbf{Variable Impedance Control (VIC) with Reinforcement Learning (RL)}. By integrating VIC with RL, the robotic system can learn and optimize its control strategies through interaction with the environment. The RL component allows the robot to explore various impedance settings, dynamically adjusting stiffness and damping. Over time, the system generalizes its learned strategies to new tasks and environments, making it highly adaptable and robust. This combination of VIC and RL enables the dual-arm system to handle complex interactions and maintain precise control, even under dynamic and uncertain conditions.
% \GN{"changing object properties and \textbf{external disturbances}" - this should be supported by some results; otherwise its better to avoid highlighting external disturbances}

\begin{figure*}[ht]
    \centering
    \captionsetup{font=footnotesize}
    \includegraphics[width=\linewidth]{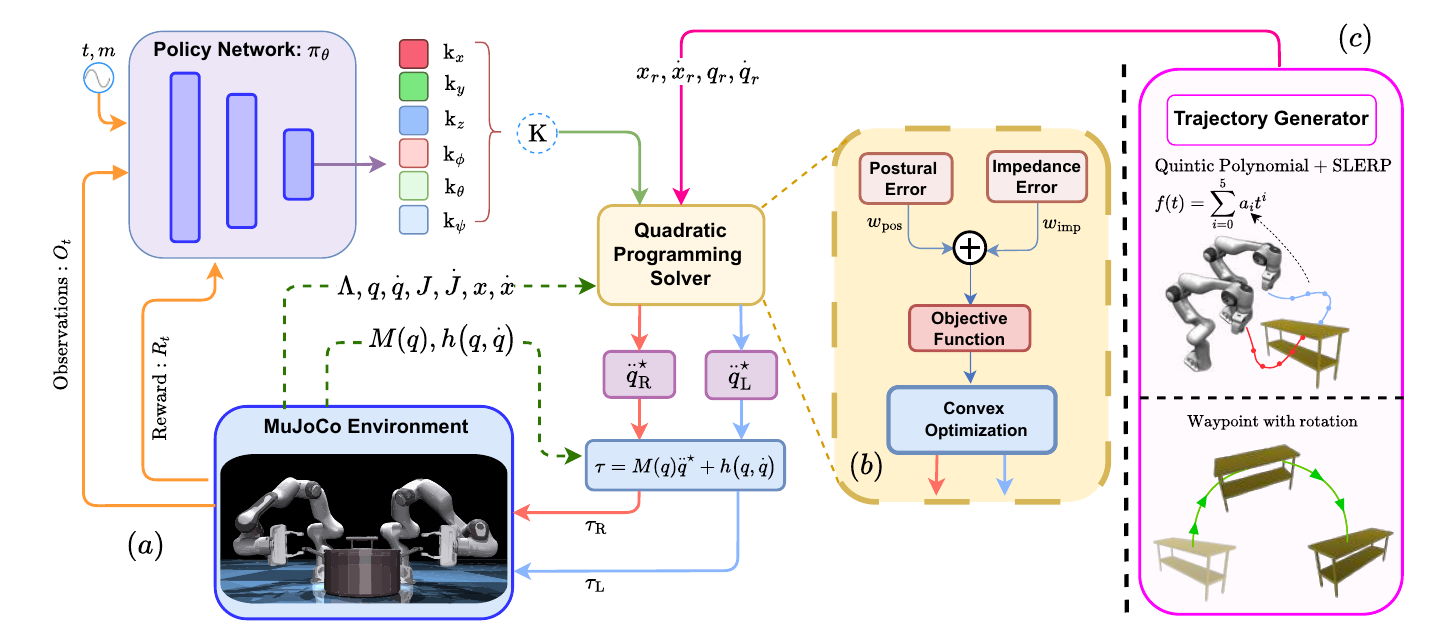}
    \caption{\textbf{Overview of the Proposed Method}: \textbf{(a)} The pipeline shows how the policy network uses observations $O_t$, reward $R_t$, time, and mass embeddings to predict stiffness $K$. This stiffness, along with state variables from MuJoCo and reference trajectory, is fed to the QP solver, which outputs joint accelerations $\boldsymbol{\ddot q^{\star}}_L$ and $\boldsymbol{\ddot q^{\star}}_R$. These accelerations are then converted to torques $\boldsymbol{\tau}_L$ and $\boldsymbol{\tau}_R$ and applied to the MuJoCo simulator. \textbf{(b)} Illustrates the QP solver implemented using CVXPY \cite{CVXPY}, which computes impedance and postural errors using the provided stiffness and reference trajectory and solves the optimization problem (Equation \ref{eq:QP_problem}) with constraints (Equations \ref{eq:constraint_1}, \ref{eq:constraint_2}, \ref{eq:constraint_3}, \ref{eq:constraint_4}) to determine the joint accelerations. \textbf{(c)} Depicts the trajectory generation process, which uses a quintic polynomial for cartesian positions and SLERP for $SO(3)$ orientations. The trajectory includes a waypoint where the $x$ and $y$ coordinates are the averages of the initial and goal positions, and the $z$ coordinate is set to a random height within the workspace of the arms to introduce variability.}
    \label{fig:pipeline}
\end{figure*}

To summarize, our contributions are:
\begin{enumerate}
    \item We propose a novel framework to perform coordinated dual-arm manipulation on multiple objects without learning from demonstrations.
    \item We demonstrate that an RL-based Variable Impedance Learning (VIL) control, implemented as a Quadratic Programming (QP), effectively adjusts robot stiffness during manipulation, enabling adaptation to varying object masses and sizes. 
    \item We demonstrate that our framework outperforms existing baselines in terms of trajectory tracking error.
\end{enumerate}

% \end{itemize}
\section{RELATED WORK}

\textbf{Dual-arm manipulation} has been addressed in various forms and task settings over the years~\cite{dualarm_survey, dualarm_traj_rl,peract2, mobile_aloha, UMI, transformer_IL, bimanual_teleop, bi_vla, deep_IL, dualarm_rl_assembly, dualarm_rl_task_adaptive, bi_touch}. 
Recent works have shown impressive results in dual-arm skill learning using imitation learning and behavioral cloning\cite{aloha, peract2, bi_vla, transformer_IL, deep_IL}. These approaches are scalable and robust to disturbances at inference time. However, imitation learning usually requires dense expert demonstrations for each task,  generally collected using teleoperation~\cite{aloha, mobile_aloha, bimanual_teleop, UMI, gello} which is a time-consuming and resource-intensive process. Moreover, adapting such policies to novel scenarios and objects without demonstrations is challenging. Recent work on achieving human-level dual-arm dexterity \cite{bidex} has shown promising outcomes in simple manipulation tasks across a wide range of objects using RL. However, these approaches are hindered by the need for extensive training and issues with generalization. While these methods emphasize learning robust skills, our work addresses the control aspect of the problem.

% \textcolor{red}{ still checking this one}

\textbf{Force coordination control of dual-arm system}\\
Force control is essential in robotics, with several key methods developed to achieve it efficiently. In master/slave control \cite{master_slave}, one arm operates under force control, adjusting to forces, while the other follows a predefined trajectory. Hybrid force/position control \cite{hybrid_force_postion_original} decomposes the task by separately managing motion and force along different axes. Impedance control \cite{impedance_control_original}, on the other hand, simultaneously regulates both motion and force, ensuring smooth dynamic interactions.
Recent methods \cite{tran_mecha_opti, bimanual_imp_2013} highlight the effectiveness of impedance control in dual-arm settings, but the stiffness (K) and damping (D) parameters are typically fixed and require manual tuning. To ensure proper interaction between the robot and environment, these parameters should adapt dynamically. VIC \cite{vic_original, abu_frontier_vic, carelli1991adaptive,li1989unified,slotine1987adaptive} addresses this need but poses challenges due to the nonlinearity of system dynamics, making parameter estimation difficult.

\textbf{Learning-based Control}
Recent advancements in torque-controlled robots have facilitated the learning of tasks that involve variable impedance. Learning controllers that generate a control action in an iterative manner to execute a prescribed action have been employed ~\cite{arimoto1984bettering}. This development enables robots to execute compliant and energy-efficient motions. 
% Many papers~\cite{adaptive_imp_control,learning_imp_icra, vices, zhang_learning_vic_irl} have used learning-based variable impedance control strategies for single arm manipulation for different tasks with different frameworks.
While learning-based VIC strategies have shown success in single-arm manipulation tasks~\cite{adaptive_imp_control, learning_imp_icra, vices, zhang_learning_vic_irl}, their direct application to dual-arm settings is limited. Dual-arm manipulation introduces challenges like coordinated force control and inter-arm dynamics, which single-arm methods cannot handle effectively.

Current dual-arm manipulation methods face limitations like the need for expert demonstrations, poor generalization, and the inefficiencies of RL. Force control often requires manual tuning of impedance parameters, while single-arm VIC methods struggle to adapt to dual-arm tasks. Our proposed framework combines RL with gradient-based optimization, enabling automatic impedance adjustments and efficient coordination of various objects across varying masses.

\section{METHOD}

Given a 3D object at an initial pose, $X_{0} \in SE(3)$ with mass $m$, a set of grasp poses $\{G_1, G_2\} \in SE(3)$ \cite{cgdf}, our goal is to control the dual-arm setup to move the object to a given final pose, $X_T \in SE(3)$, where $T$ refers to the total number of rollout steps. Our task is formulated as a trajectory-tracking problem, where the dual arms, follow two end-effector trajectories while maintaining coordination in manipulating the object. 
We first generate an object trajectory by interpolating points between the initial and the final pose. 
The end effectors are constrained to remain stationary w.r.t the object at all times. Thus, we use the object trajectory to compute individual end-effector trajectories.
In order to maintain the compliance needed to safely actuate in this setting, the interaction between the object and the end effectors is modelled as a mass-spring-damper system via Impedance Control.

% \GN{Some information about impedance controller should be included here: The interaction between the object and the
% end-effector is modeled as a mass-spring-damper system.}(done)

As shown in Fig.~\ref{fig:pipeline}, our proposed method has two components: (1) A \textbf{policy network} that predicts the stiffness parameters \(K\), and (2) A \textbf{convex optimization-based controller} that uses the stiffness (\(K\)) parameters to calculate the joint accelerations (\(\boldsymbol{\ddot{q}}\)), and subsequently, torques (\(\boldsymbol{\tau}\)) which are executed in simulation. The damping parameters \(D\), are set such that the system is always critically damped. 

In the following subsections, we explain the formulation of each component of \acro. 
\subsection{Policy Network} 
We implement our policy network $\pi_{\theta}$ using the popular \textbf{Proximal Policy Optimization (PPO)} algorithm \cite{ppo}. Our \textbf{observation space} at a timestep $t$ consists of the change in end-effector pose of left and right arms $\{\Delta \boldsymbol{x}_L, \Delta \boldsymbol{x}_R\} \in SE(3)$, the object pose $X_t \in SE(3)$, joint positions of both 7-dof arms $\{\boldsymbol{q_L}, \boldsymbol{q_R}\} \in \mathbb{R}^7$, end-effector wrench for both arms $\{\boldsymbol{f}_L, \boldsymbol{f}_R\} \in \mathbb{R}^6$, previous actions $K_{t-1} \in \mathbb{R}^{6}$, and sinusoidal embeddings for timestep and mass, both $\in \mathbb{R}^8$~\cite{vaswani2017attention}.
We input each of the $SE(3)$ elements as a vector of length 7 consisting of cartesian coordinates for position and quaternion for rotation. 
This observation space is processed through a feature extractor, implemented as a 4-layer MLP. The policy network and the value functions are modelled as a 5-layer MLP. 
The \textbf{action space} is defined as a set of discrete action bins in multiples of 20, predicting the 6 diagonal stiffness parameters $K = \text{Diag}\left[k_x, k_y, k_z, k_\phi, k_\theta, k_\psi\right]$. Using a classification objective adds stability to the training and provides well-defined decision boundaries between different values of \(K\). The inclusion of joint positions and wrench in the observation space helps the model better understand the dynamics of the simulation environment. 

% \GN{Typos - 1) Line4: end-effector's poses not positions, 2) forces/moments or wrench $\in R^6$, 3) both $\in R^8$ not clear, 4) $\Delta g$ is not used anywhere else, 5) dimension of K - is it $R^6$ or $R^{6x6}?$, 6) the grasp poses should be explained somewhere} (done)

% \GN{Just pointing out the typos: 1) ensure that are we considering only $f_ext1 \in \mathbb{R}^3 $ or forces and moments (wrench space $\in \mathbb{R}^6$), 2) There are two $K_y$s in the stiffness parameters- may be subscripts $\phi,\theta, \psi$ can be used 3) L and R or 1 and 2 for representing two arms?? - in the pipeline they were $\ddot{q}_left$ and right 4) notations should be consistent throughout - scalars: lowercase or upper, vectors: bold lowercase, matrix: bold upper case \\ also the overview of pipeline should have some more technical details}

% rewrite with specifics
\textbf{Rewards:} We provide dense rewards throughout the episode to promote desirable behaviors and discourage undesirable ones. Positive rewards are given for minimizing the error in the end-effector pose and object pose given by the reference trajectory, as well as for successfully moving the object to the goal position. A negative reward is applied if the predicted stiffness results in an infeasible solution by the optimizer. To ensure smooth stiffness changes, we also track an Exponential Moving Average (EMA) of the predicted \(K\) values. If the current stiffness prediction deviates from the EMA by more than a set threshold, a negative reward is given. This helps maintain consistency in \(K\) adjustments over time. Additionally, smooth \(K\) values are essential as significant variations in stiffness can cause abrupt torque changes, leading to sudden forces on the robotic arms~\cite{ema_issue}. This can strain motors and other mechanical components, potentially causing damage. We show the effectiveness of EMA as an ablation in the section~\ref{sec:RESULTS}.

The $K$ predicted by the policy network is then passed onto the optimization-based controller which solves for $\boldsymbol{\ddot{q}}$, which is then used to compute the $\boldsymbol{\tau}$ using the system dynamics (\ref{system_dynamics}). These torques are given to the manipulator as control inputs.
\begin{table}[h]
\renewcommand{\arraystretch}{1.3}
\centering
\captionsetup{font=footnotesize}
\begin{tabular}{|c|l|}
\hline
\textbf{Symbol} & \textbf{Description} \\ \hline
$\boldsymbol{x}$, $\boldsymbol{\dot{x}}$, $\boldsymbol{\ddot{x}}$ & end effector pose, twist and acceleration respectively \\ \hline
$\boldsymbol{q}$, $\boldsymbol{\dot{q}}$, $\boldsymbol{\ddot{q}}$ & joint positions, velocities and accelerations respectively  \\ \hline
$\boldsymbol{f}$            & Wrench \\ \hline
$M(\boldsymbol{q})$         & Mass matrix in Joint Space \\ \hline
$J(\boldsymbol{q})$, $\dot{J}(\boldsymbol{q})$  & Jacobian and its derivative \\ \hline
$\Lambda(\boldsymbol{q})$   & Inertia matrix in Task Space \\ \hline
$\boldsymbol{h}(\boldsymbol{q},\boldsymbol{\dot{q}}) $ & Centripetal, Coriolis and Gravitational forces\\ \hline
$K$            & Stiffness Matrix (\(\mathbb{R}^{6 {\times} 6}\)) \\ \hline
$D$            & Damping Matrix (\(\mathbb{R}^{6 {\times} 6}\)) \\ \hline
%  $\boldsymbol{e}_{\text{imp}}$ & Error in accelerations in end-effector \\ & space for impedance task \\ \hline
% $\boldsymbol{e}_{\text{pos}}$ & Error in accelerations in joint space \\ & for posture task \\ \hline
% $w_{\text{imp}}$ & weight of the impedance task \\ \hline
% $w_{\text{pos}}$ & weight of the postural task \\ \hline
\end{tabular}
\caption{Symbols used and their descriptions. Throughout our notation, vectors have been represented in \textbf{bold} and matrices are shown in \textit{Capitalised Italics}. The subscripts $L$ and $R$
refer to the left and right manipulators respectively. The \textit{absence} of subscripts $L$ and $R$ implies the quantity is being considered for both the manipulators i.e. $\boldsymbol{x} \text{ denotes both }  \boldsymbol{x}_L \text{ and } \boldsymbol{x}_R$}
\label{tab:symbols}
\end{table}

\begin{figure*}[ht]
    \centering
    \captionsetup{font=footnotesize}
    \includegraphics[width=1\linewidth]
    {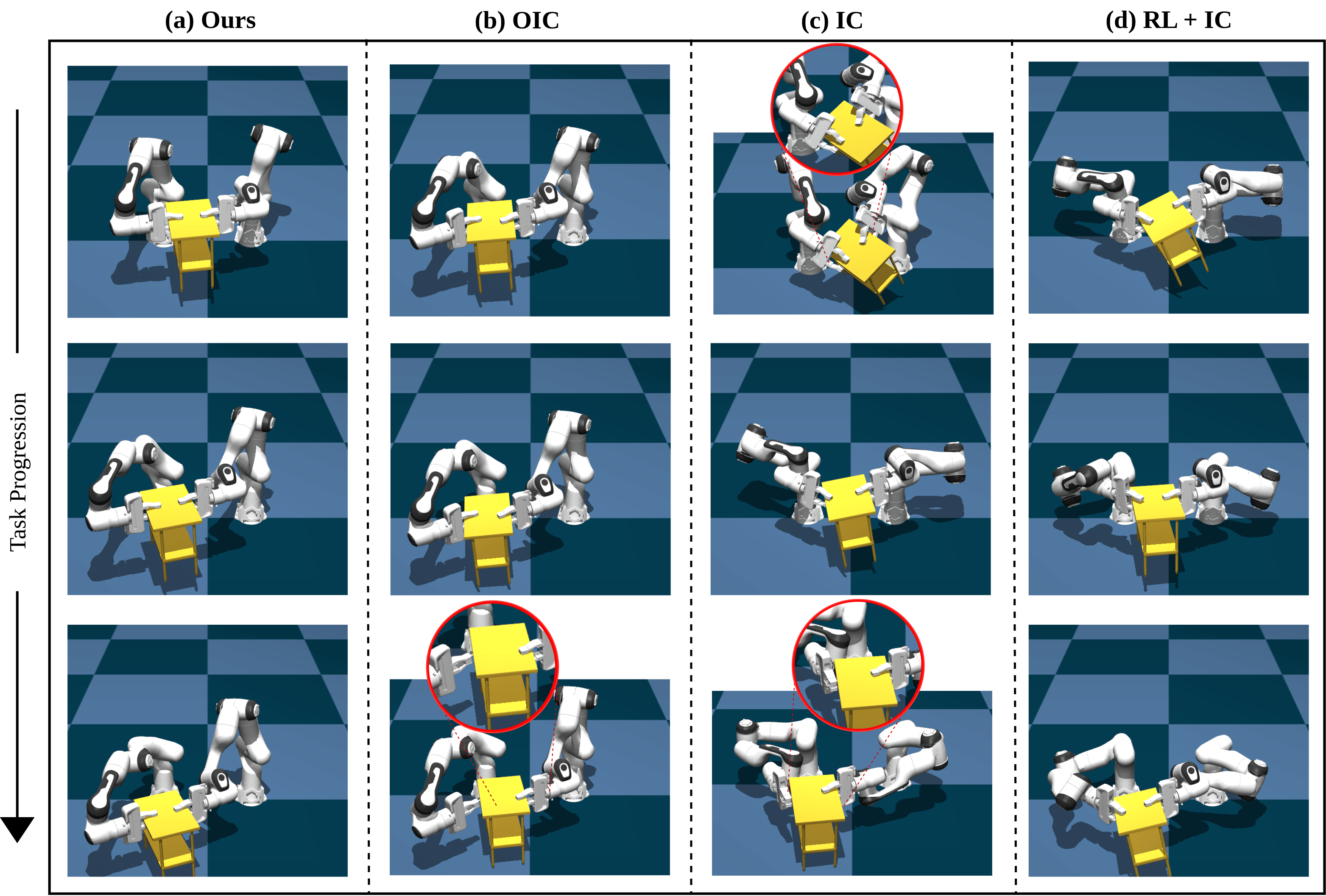}
    \caption{Qualitative comparison of our approach with baseline methods: \textit{Optimization-based Impedance Control (OIC)}, \textit{Impedance Control(IC)}, and \textit{RL-based Impedance Control (RL+IC)}. (\textbf{a}) Our framework successfully completes the pick-and-place task with the stool, while (\textbf{b}) Optimization-based Impedance Control (OIC) achieves similar results but exhibits object slipping (zoomed in the red circle) when the arms are fully extended. (\textbf{c}) Impedance Control (IC) fails to complete the task due to its inability to adapt impedance parameters dynamically, leading to poor object handling. (\textbf{d}) RL-based Impedance Control (RL + IC)  completes the task without slipping but deviates from the reference trajectory mid-task.}
    \label{fig:qualitative_results}
\end{figure*}
\subsection{Optimization-Based Controller}
Our goal is to achieve trajectory tracking of the object moving from $X_0$ to $X_T$ while ensuring all physical constraints are satisfied. The system must also account for the interaction forces that arise during manipulation. To address these challenges, we  adapted the Quadratic Programming (QP) - based controller from \cite{QP_for_dualarm}, customizing it for our \acro framework for dual-arm manipulation with two-fingered gripper end-effectors.

The \(K\) values predicted by the policy network are used to calculate the joint accelerations (\(\boldsymbol{\ddot{q}}\)) necessary for tracking the desired trajectory (\(\boldsymbol{x_r}\)). We use CVXPY \cite{CVXPY} for solving the QP, which determines the joint accelerations by minimizing the tracking error while accounting for the interaction forces. However, with a redundant 7-DOF manipulator, multiple solutions may exist, allowing for the execution of various tasks concurrently by leveraging this redundancy. We consider two specific tasks: 1) Impedance Task and 2) Postural Task. The rationale for selecting these tasks is explained subsequently.

\textbf{Impedance Task :} This task aims to minimize tracking errors while considering the interaction forces. The impedance model primarily depends on the spring and damper components, which govern the interaction forces. The wrench $f$ is modelled as, 
\begin{gather}
    \label{interaction_forces}
    \boldsymbol{f}  = D(\dot{\boldsymbol{x}_r} - \dot{\boldsymbol{x}}) + K(\boldsymbol{x}_r - \boldsymbol{x})
\end{gather}

 The stiffness matrix \(K\) and the damping matrix \(D\) are computed using the expression \(\sqrt{\Lambda}\sqrt{K} + \sqrt{K}\sqrt{\Lambda}\) and these matrices are incorporated into the cost function for this task. The cost function of the QP for this task is expressed as:
\begin{gather}
    \boldsymbol{e}_{\text{imp}} = \boldsymbol{\ddot{x}}- \Lambda^{-1} \boldsymbol{f}
\end{gather}
where,\hspace{23mm} \(\Lambda : = (J M^{-1}J^{T})^{-1}\)
\begin{gather}
         \boldsymbol{\ddot{x}} = J \boldsymbol{\ddot{q}} + \dot{J}\boldsymbol{\dot{q}}
\end{gather}
\textbf{Posture Task :} To maintain the subsequent joint configurations of the dual-arm system close to the configurations established during contact with the object, denoted as $\boldsymbol{q_r}$, the following postural task is implemented.
\begin{gather}
    \boldsymbol{e}_{\text{pos}} = \boldsymbol{\ddot{q}} - 2\sqrt{K_{null}}\ (\boldsymbol{\dot{q}}_{r} - \boldsymbol{\dot{q}}) + K_{null}\ (\boldsymbol{q}_{r} - \boldsymbol{q})
    \label{eq:postural_task}
\end{gather}
where \(K_{null}\) is the null-space proportional gain. \\

The \textbf{full QP formulation} is given by :
\begin{gather}
    \left(\boldsymbol{\ddot{q}}_L^*, \boldsymbol{\ddot{q}}_R^*\right) = \arg\min\limits_{\left(\boldsymbol{\ddot{q}}_L, \boldsymbol{\ddot{q}}_R\right)}  w_{\text{imp}} \left(\|\boldsymbol{e}_{\text{imp}_{L}}\|^2_2 + \|\boldsymbol{e}_{\text{imp}_{R}}\|^2_2\right) \notag \\
    % \right\rVert^2 + w_{\text{pos}} \lVert e_{\text{pos}_L}\rVert^2 
    + w_{\text{pos}} \left( \|\boldsymbol{e}_{\text{pos}_L}\|^2_2 + \|\boldsymbol{e}_{\text{pos}_R}\|^2_2 \right)
    \label{eq:QP_problem}
\end{gather}
\setlength{\belowcaptionskip}{2pt}
\begin{figure}[h]
    \centering
    \captionsetup{font=footnotesize}
    \includegraphics[width=1\linewidth]{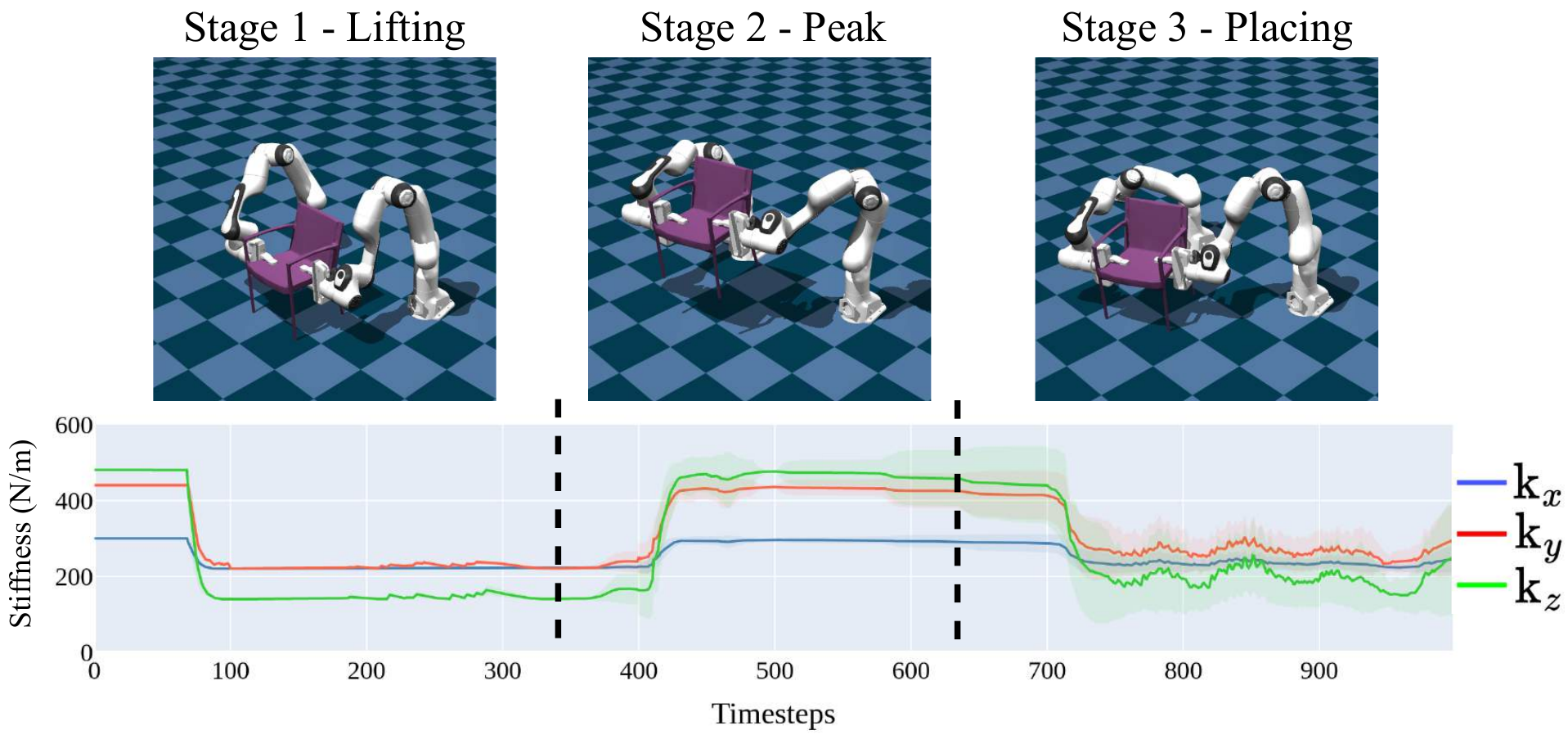}
    \caption{Stiffness ($K$) values during pick-and-place of the chair (5kg). In \textbf{Stage 1}, $K$ values are low at motion initiation. \textbf{Stage 2} shows an increase in $K$ to reach the intermediate waypoint. \textbf{Stage 3} sees $K$ return to initial levels during object placement.}
    \label{fig:kplot}
\end{figure}

subject to the constraints,
\begin{gather}
    \boldsymbol{q_{\text{min}}} \leq \frac{1}{2} \boldsymbol{\ddot{q}} \Delta t^2 + \boldsymbol{\dot{q}} \Delta t + \boldsymbol{q} \leq \boldsymbol{q_{\text{max}}} 
    \label{eq:constraint_1}\\
    \boldsymbol{\dot{q}_{\text{min}}} \leq \boldsymbol{\ddot{q}} \Delta t + \boldsymbol{\dot{q}} \leq \boldsymbol{\dot{q}_{\text{max}}} 
    \label{eq:constraint_2}\\
    \boldsymbol{\tau_{\text{min}}} \leq M \boldsymbol{\ddot{q}} + \boldsymbol{h} \leq \boldsymbol{\tau_{\text{max}}}
    \label{eq:constraint_3}\\
    \| \left(\frac{1}{2}J_L\boldsymbol{\ddot{q}}_L\Delta t^2 + J_L\boldsymbol{\dot{q}}_L\Delta t + \boldsymbol{x}_L \right) - \notag \\
    \left(\frac{1}{2}J_R\boldsymbol{\ddot{q}}_R\Delta t^2 +J_R\boldsymbol{\dot{q}}_R\Delta t + \boldsymbol{x}_R \right) \|_2 \leq W_G + \textit{tol}
    \label{eq:constraint_4}
\end{gather}

The optimizer ensures that the robot's joint accelerations $\boldsymbol{\ddot{q}}$ and torques $\boldsymbol{\tau}$ satisfy physical constraints such as joint limits, velocity limits, and torque limits which have been taken from the robot's datasheet \footnote{\href{https://download.franka.de/documents/100010_Product Manual Franka Emika Robot_10.21_EN.pdf}{Franka Emika Robot's Instruction Handbook}}.
Furthermore, equation~\ref {eq:constraint_4} is used to constrain the distance between the end effectors to match the distance between the grasps $(W_G)$ with some error tolerance $(\textit{tol})$.
% This is beneficial for proper coupling between both arms.
Using the joint accelerations, the torques are computed as 
\begin{gather}
    \label{system_dynamics}
    \boldsymbol{\tau} = M(\boldsymbol{q})\boldsymbol{\ddot{q}^*} + \boldsymbol{h}(\boldsymbol{q},\boldsymbol{\dot{q}}).
\end{gather}
We provide the torques to the simulation environment, which then updates its state. Finally, we compute the tracking error based on the updated state variables.

\section{EXPERIMENTS}
\label{sec:EXPERIMENTS}
In this section, we evaluate our proposed framework on large, complex objects with varying masses and across different goal positions and trajectories for the coordinated pick-and-place task in a dual-arm setting. 
% \GN{various trajectories or different goal poses are not considered?}

% \begin{figure}[h]
%     \centering
%     \includegraphics[width=0.75\linewidth]{images/Rendering.jpeg}
%     \caption{Pick and Place Task in Mujoco Environment}
%     \label{fig:task_render}
% \end{figure}

\textbf{Task Setup:} We use the MuJoCo simulator \cite{mujoco} for all our experiments. As shown in Fig.~\ref{fig:enter-label}, the pick-and-place task is conducted in the simulation to evaluate the proposed approach. In our experiments, we choose six objects --- \textit{chair, stool, stockpot, laptop, monitor} and \textit{crate}. Each object presents unique challenges due to differences in size and shape. For example, the chair and stool are large objects and require careful coordination between the dual arms for balance. \begin{figure}[h]
    \centering
    \captionsetup{font=footnotesize}
    \includegraphics[width=1\linewidth]
    {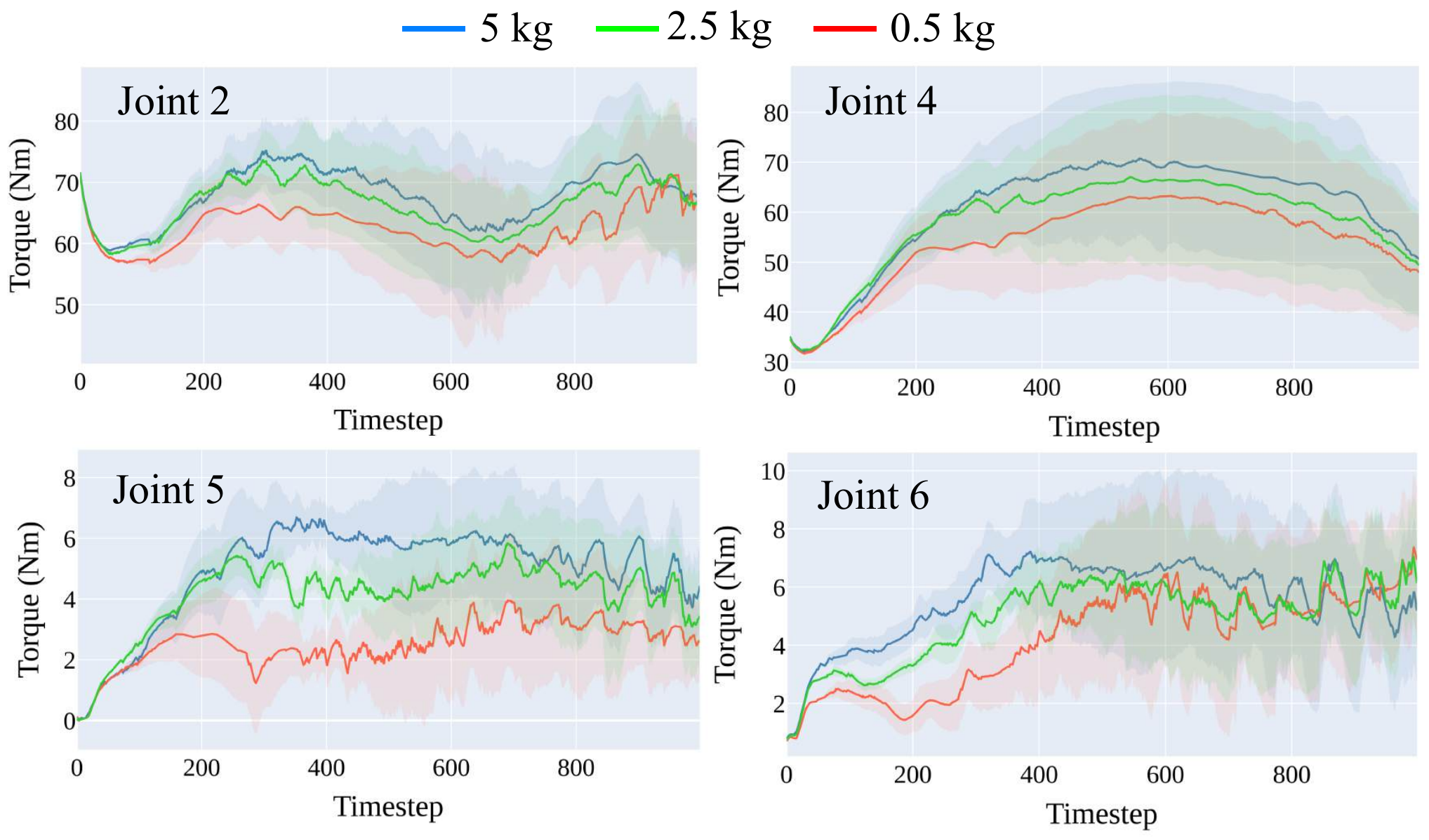}
    \caption{Torque values of different joints from our method for a pick and place task with three different masses (5kg, 2.5kg, 0.5kg). The torque values increase for higher masses, while smaller masses result in lower torque values (Joint notation indexing starts from Joint 1).}
    \label{fig:quantitative_torque}
\end{figure}Meanwhile, the stockpot has an irregular shape and constraints the arms to grasp it using the handles. Additionally, to introduce variability in the learning process, we vary the mass of the objects among 0.5, 1, 2.5, and 5kg. 

% \Madhav{Is there any other task we are doing other than pick and place. The sentence below conveys that. We can have a section on Assumptions wherein we state all assumptions and where we can say we assume that the end effector poses are given by a method such as CGDF or manually. If this is the only given we can just write it in a single sentence that the grasps are an input to the system }
% In the pick-and-place task, we assume that the grasp poses are known and stable. A reference trajectory is generated using the quintic polynomial for each axis $(x, y \ \text{and}\ z)$ for the desired waypoints. Additionally, we use SLERP interpolation for interpolating between the initial and the goal orientation.
As illustrated in Fig.~\ref{fig:enter-label}, the pick-and-place task is executed in three stages: grasp, pick and place. We define the initial and goal positions of the object, with a waypoint for the second stage positioned at the midpoint along the $x$ and $y$ axes and a randomly specified value along the $z$ axis to simulate the pick-and-place operation. To determine the trajectories for these three stages, we employ quintic polynomial and spherical linear interpolation (SLERP). The task involves multiple grasps per object, which are assumed to be known and stable. Throughout the MuJoCo simulation, the environment computes the dynamic model parameters such as $M(\boldsymbol{q})$ and $\boldsymbol{h}(\boldsymbol{q},\boldsymbol{\dot{q}}) $, to accurately represent the robot's behavior during the task execution. 
\textbf{Baselines: }Existing RL and MPC-based impedance control approaches~\cite{vices, mpvic, MPC_VIC_HCI}  are not easily transferable to dual-arm setups due to their inability to handle dual-arm constraints and interaction forces. Furthermore, the absence of publicly available RL training codes for these works complicates direct comparisons. Additionally, imitation learning and behavioral cloning methods \cite{mobile_aloha, bi_vla, UMI, bi_touch, deep_IL} need expert demonstrations to train, thus making a comparison challenging, as we do not require such demonstrations. To evaluate the performance of our framework and ensure a fair comparison, we implemented and included the following baselines:

\textit{Optimization-based Impedance Control (OIC)}: Following \cite{QP_for_dualarm}, we formulate the pick-and-place task using a conventional impedance controller with a QP formulation. For this method, we provide arbitrary stiffness values $(K)$ uniformly sampled from a range identified through our method. (Refer to the 2nd column of Table~\ref{tab:object_comparison})
% \Madhav{Please mention here which column of the table this method corresponds to. Also mention in the captions as well. Let all captions be clear and self contained}. 

\textit{Reinforcement Learning with Impedance Control (RL+IC)}: In this baseline, RL predicts the stiffness values, which are then used by a conventional impedance control for task execution. The stiffness values are dynamic and based on the policy's predictions. (Refer to the 4th column of Table~\ref{tab:object_comparison})
\begingroup
\begin{table}[ht]
    \centering
    \captionsetup{font=footnotesize}
    \renewcommand{\arraystretch}{1.3}
    \adjustbox{max width=\textwidth}{
    \begin{tabular}{|c|c|c|c|c|c|}
        \hline
        Object & \textbf{Ours} & OIC & IC & RL+IC & Ours \\
        & & & & & w/o EMA \\
        \hline
        Chair & \textbf{0.0131} & 0.0166 & 0.0598 & 0.0612 & 0.0227\\
        Monitor & \textbf{0.0428} & 0.0712 & 0.1413 & 0.1085 & 0.1189\\
        Laptop & \textbf{0.0406} & 0.0555 & 0.1564 & 0.0696 & 0.0721\\
        Stockpot & \textbf{0.0277} & 0.0384 & 0.2201 & 0.0642 & 0.1153\\
        Stool & \textbf{0.0172} & 0.2547 & 0.1451 & 0.1074 & 0.0444\\
        Crate & \textbf{0.0310} & 0.0378 & 0.1415 & 0.0760 & 0.0950\\
        \hline
        \multicolumn{1}{|c|}{\textbf{All objects}} & & & & & \\
        \multicolumn{1}{|c|}{\textbf{(1kg)}} & \textbf{0.0367} & 0.0554 & 0.1375 & 0.0835 & 0.0780 \\
        \hline
    \end{tabular}}
    \caption{Mean tracking errors (in meters) of different methods across all objects and masses.}
    \label{tab:object_comparison}
\end{table}
\endgroup

\textit{Impedance Control (IC)}: This baseline is a conventional impedance controller 
\cite{impedance_control_original} with no Optimization or RL as an aid to it. (Refer to the 3rd column of Table~\ref{tab:object_comparison})

\textbf{Metrics:} We use object tracking error as our metric, defined as the mean tracking error of the object across all trajectories and object masses. It provides a clear measure of how accurately the arms follow the reference trajectory during manipulation, with lower errors indicating better control and secure handling of the object without loss of grip.  
We define the tracking error as follows, 
\begin{gather}
e_{\text{tracking}} = \frac{1}{|\mathcal{M}|} \sum_{m \in \mathcal{M}} \frac{1}{T} \sum_{t = 1}^{T} \|\boldsymbol{x_{\text{ref}}}(t) - \boldsymbol{x_{\text{actual}}}(t) \|_2
\end{gather}
where $ \mathcal{M} = \{0.5, 1, 2.5, 5\}$ is the set of all masses (in kg).
We evaluate each model on 100 uniformly sampled goal locations across the workspace for every object mass and report the above metric. 
% \Madhav{Last column of the table is empty. Also a lot of space between the table and RESULTS}
\section{RESULTS}
\label{sec:RESULTS}
\begin{figure}[ht]
    \centering
    \captionsetup{font=footnotesize}
    \includegraphics[width=\linewidth]{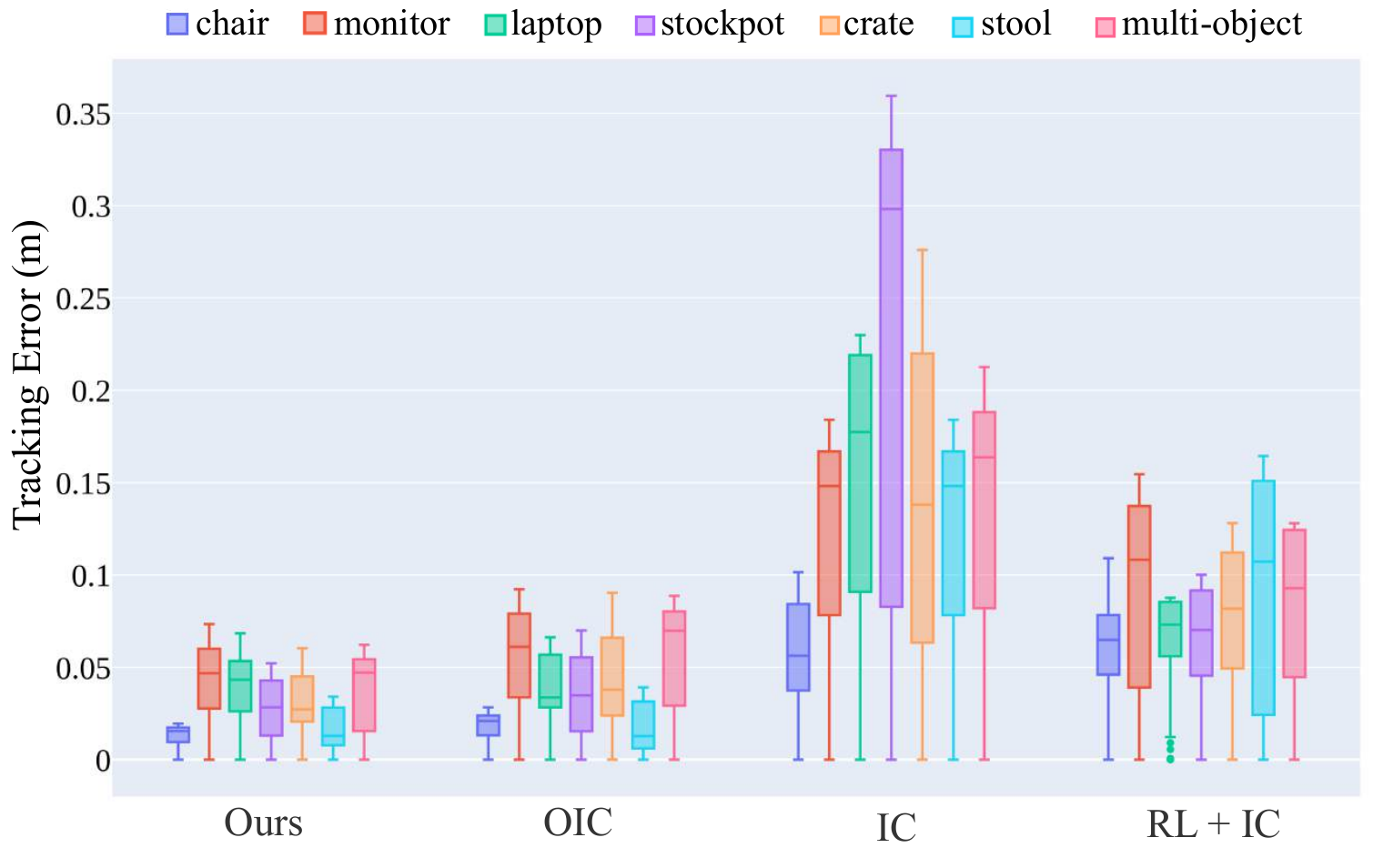}
    \caption{Variance in tracking errors across different methods and object masses: We compute the variance in mean tracking error of each method across 100 trajectories for all masses (as detailed in section~\ref{sec:EXPERIMENTS}). For the multi-object case, the object's mass was set to 1 kg.}
    \label{fig:quantitative_tracking_error}
\end{figure}

As shown in Table~\ref{tab:object_comparison} and Fig.~\ref{fig:quantitative_tracking_error}, our method consistently outperforms all baselines in terms of trajectory tracking error for all objects across all masses defined in section~\ref{sec:EXPERIMENTS}. The final row of Table~\ref{tab:object_comparison} shows results for a single model trained on all objects of 1kg, where our method still demonstrates superior performance compared to the baselines.
From Fig.~\ref{fig:quantitative_tracking_error}, we observe lower variance in tracking error over 100 trajectories, indicating improved accuracy where other methods either fail or exhibit higher errors.

Figure~\ref{fig:qualitative_results} illustrates that our method completes the task successfully and is able to place the object easily at a distant goal pose. Moreover, Impedance Control and RL-based Impedance Control fail to complete the task as they face issues with object slipping and deviations from the desired trajectory. Although Optimization-based Impedance Control performs similarly, it struggles with distant goal locations due to its limited ability to adjust impedance parameters. 

In Fig.~\ref{fig:kplot}, the \(K\) values increase as the arms reach the intermediate waypoint, where an extended configuration is required. The adaptive stiffness facilitates this transition, leading to successful task completion and lower tracking errors.
% In Fig.~\ref{fig:kplot}, we observe that the \(K\) values increase while reaching the intermediate waypoint, which requires the arms to have an extended configuration. To comply with this, adaptive stiffness is required and has been observed which helps in successful task completion resulting in lower tracking errors.
% As observed in Fig.~\ref{fig:kplot}, having an adaptive stiffness throughout the episode helps in the successful completion of the task, hence resulting in lower tracking errors.
% This also explains the general increasing trend in joint torques (\(\boldsymbol{\tau}\)). Fig \ref{fig:quantitative2} displays the joints bearing the largest load in our task. 
% Naturally, in order to maintain minimum tracking error, the $\boldsymbol{\tau}$ values increase with increasing mass.
We also observe that the torques required for joints that are primarily involved in pick-and-place tasks (i.e. joints 2, 4, 5, 6) increase with increasing object mass, as shown in Fig.~\ref{fig:quantitative_torque}. This shows that our method adapts to these mass variations effectively. 

Finally, we performed an ablation study on the reward function by removing the EMA reward component and training models for pick-and-place tasks with objects of varying mass. The increase in tracking error is shown in the last column of Table~\ref{tab:object_comparison}. A significant rise in tracking error was observed for larger objects such as the monitor, stockpot, and crate. For these objects, it is critical that the predicted stiffness (K) values remain smooth throughout the time to avoid abrupt changes in torque, which would complicate the overall handling. This emphasizes the necessity of stabilizing action predictions for effective control.

\section{CONCLUSIONS}
In this work, we tackle the challenges associated with coordinated dual-arm manipulation, which requires precise control strategies due to the dynamic interactions between two arms and objects. Traditional methods, such as Impedance Control, often struggle with fixed parameter tuning, making them inadequate for tasks involving varying object properties. To address these issues, we propose an integrated framework that combines Reinforcement Learning (RL) with an optimization-based Variable Impedance Control (VIC), allowing the system to adjust stiffness and damping, enabling it to handle objects of varied mass and shape. The system's adaptability demonstrates potential for real-world applications that require dynamic task handling, especially in scenarios with unpredictable object characteristics. We evaluate our framework on existing control methods extended for a dual-arm setting and show its effectiveness through enhanced performance in dual-arm pick-and-place task for six objects of varying size and mass. 

Future work may involve integrating new complex tasks, such as assembly or collaborative manipulation, incorporating collision-avoidance-based trajectory generation to navigate environments with obstacles, and testing the framework on real hardware to address challenges such as sensor noise and physical dynamics.

\bibliographystyle{IEEEtran}

\bibliography{bibtex}

\end{document}